# Predicting Early Dropout: Calibration and Algorithmic Fairness Considerations

**Marzieh Karimi-Haghighi**
Universitat Pompeu Fabra
marzieh.karimihaghighi@upf.edu

**Carlos Castillo**
Universitat Pompeu Fabra
carlos.castillo@upf.edu

**Davinia Hernández-Leo**
Universitat Pompeu Fabra
davinia.hernandez-leo@upf.edu

**Veronica Moreno Oliver**
Universitat Pompeu Fabra
veronica.moreno@upf.edu

**ABSTRACT**: In this work, the problem of predicting dropout risk in undergraduate studies is addressed from a perspective of algorithmic fairness. We develop a machine learning method to predict the risks of university dropout and underperformance. The objective is to understand if such a system can identify students at risk while avoiding potential discriminatory biases. When modeling both risks, we obtain prediction models with an Area Under the ROC Curve (AUC) of 0.77-0.78 based on the data available at the enrollment time, before the first year of studies starts. This data includes the students' demographics, the high school they attended, and their admission (average) grade. Our models are calibrated: they produce estimated probabilities for each risk, not mere scores. We analyze if this method leads to discriminatory outcomes for some sensitive groups in terms of prediction accuracy (AUC) and error rates (Generalized False Positive Rate, GFPR, or Generalized False Negative Rate, GFNR). The models exhibit some equity in terms of AUC and GFNR along groups. The similar GFNR means a similar probability of failing to detect risk for students who drop out. The disparities in GFPR are addressed through a mitigation process that does not affect the calibration of the model.

## 1    INTRODUCTION

About 36% of university students in the European Union, 39% in the US, 20% in the Australia and New Zealand, and 52% in Brazil discontinue their studies before graduation (Vossensteyn, 2015; Shapiro, 2017; OECD, 2016). Reducing the rate of dropout and underperformance is crucial as these lead to social and financial losses. In addition, detecting students at risk as early as possible is necessary to improve learning and prevent them from quitting and failing their studies.

Research on actionable indicators that can lead to interventions to reduce dropout has received increased attention in the last decade, especially in the Learning Analytics (LA) field (Siemens, 2013; Viberg, 2018; Sclater, 2017; Leitner, 2017). These indicators can help provide effective prevention strategies and personalized intervention actions (Romero, 2019; Larrabee Sønderlund, 2019). Machine Learning (ML) methods, which identify patterns and associations between input variables and the predicted target (Pal, 2012), have been shown to be effective at this predictive task in many LA studies (Plagge, 2013; Kemper, 2020; Aulck, 2016; Nagy, 2018; Del Bonifro, 2020).

We remark that among students who discontinue their studies, some sub-groups are over-represented, something that needs to be considered when developing ML methods. For example, in the UK elder students at point of entry (over 21 years) are more likely to drop out after the first year compared to younger students who enter university directly from high school (Larrabee Sønderlund, 2019). In the US, graduation rate among ethnic minority university students is lower than among White students (Shapiro, 2017). Disparities in risks have been studied in previous work (Gardner, 2019; Hutt, 2019; Kizilcec, 2020) and are addressed in our work by performing per-group analysis of dropout risk and algorithmic bias mitigation of the risk predictions across different groups.

**Our contribution**. We observe a high dropout rate (43%) among computer engineering undergraduate students at a university in a multinational country in Europe[1] – in comparison to the average EU university students' dropout rate (36%) (Vossensteyn, 2015). In this work, we predict the risk of university dropout and underperformance in this engineering school. Calibrated ML models, having outputs that can be directly interpreted as probabilities for dropout or underperformance, are created using student's features available at the time of enrolment (before students start their studies). It is notable that dropout can also be due to the lack of some qualitative variables in the engineering field, such as motivation or vocational changes (Salas-Morera, 2019) in addition to the institutional rules. We evaluate our models for accuracy and fairness, as model learning may lead to unfairness for some sensitive groups (Corbett-Davies, 2018; Chouldechova A. a., 2018; Barocas, 2017; Mehrabi, 2019; Zou, 2018). Some of the disparities found are addressed through a mitigation procedure (Pleiss, 2017), which seeks to equalize error rates (generalized false positive rate or generalized false negative rate) across groups while preserving the calibration in each group.

The rest of this paper is organized as follows. Section 2 outlines related work. In Section 3, the dataset used in this study is described. The methodology including the ML models and algorithmic fairness analysis are presented in Section 4. Results are given in Section 5, and a procedure to mitigate algorithmic discrimination is used in Section 6. Finally, conclusions and recommendations are presented in Section 7.

## 2  RELATED WORK

Machine Learning (ML) methods have been used to predict dropout in higher education. In a paper (Aulck, 2016), the impact of ML on undergraduate student retention is investigated by predicting students dropout (defined as not completing at least one undergraduate degree within 6 calendar years of first enrollment). Using students' demographics and academic transcripts, different ML models result in AUCs between 0.66 and 0.73. In another study (Nagy, 2018), an early university dropout is predicted based on available data at the time of enrollment (personal data and secondary school performance) using several ML models with AUCs from 0.62 to 0.81. Similarly, in a recent study (Del Bonifro, 2020), several ML methods are used to predict the dropout of first-year undergraduate students before the student starts the course or during the first year.

Several studies (Chouldechova A. a., 2018; Corbett-Davies, 2018; Barocas, 2017; Mehrabi, 2019; Zou, 2018), have shown that ML models may lead to discriminatory outcomes for some sensitive groups.

---

[1] Country and university name omitted in this version for double-blind review.

There are many different definitions of algorithmic fairness (Narayanan, 2018), some of which are incompatible with one another. It is impossible to satisfy all of them simultaneously except in pathological cases (such as a perfect classifier), and in general it is impossible to maximize algorithmic fairness and accuracy at the same time (Berk R. , 2019). Hence, there are necessary trade-offs between different metrics (Kleinberg, 2016). Some studies (Hardt, 2016; Zafar, 2017; Woodworth, 2017) try to mitigate potential algorithmic discrimination by introducing a penalization term for unfairness in an objective function to be optimized. Also, several studies (Zemel, 2013; Kamiran, 2009; Kamishima, 2011) tried to approach statistical parity in which the same probability of receiving a positive-class prediction is considered for different groups.

One of the closest studies to ours (Gardner, 2019), considers algorithmic fairness of predictive models of students dropout in MOOCs in terms of accuracy equity using the Absolute Between-ROC Area (ABROCA) metric. The method to improve algorithmic fairness is slicing analysis, which is also used in another study (Hutt, 2019) to analyze fairness across sociodemographic groups in a predictive ML modeling of on-time college graduation. In comparison, in this study we create calibrated ML models that can predict dropout and underperformance risks solely from information available at the time of enrollment, and that have passed through a bias mitigation procedure to avoid error disparities while keeping calibration.

Calibration means that the output of the classifier is not merely a score, but an estimate of the probability of the (adverse) outcome. When we talk about fairness across two groups, we would like this calibration condition to hold for the cases within each of these groups as well. Due to the importance of calibration in risk assessment tools (Berk R. a., 2018; Dieterich, 2016), some previous work has tried to minimize error disparity across groups while maintaining calibration (Pleiss, 2017). In Pleiss et al.'s work, which is closely related to ours but for a different domain, algorithmic bias in a machine learned risk assessment task (criminal recidivism) is minimized by equalizing generalized false positive rates along different racial backgrounds, finding this equalization to be incompatible with calibration. In contrast, in the work presented on this paper, we try to minimize bias in dropout predictive ML models by equalizing error rates (generalized false positive rate or generalized false negative rate) along some sensitive groups while preserving calibration in each group. Finally, we find that equalization along some groups is not entirely incompatible with calibration.

## 3   DATASET

The anonymized dataset used in this research have been provided by a university in a multinational country in Europe and consists of 881 computer engineering undergraduate students who first enrolled between 2009 and 2017. From this population, 31 cases who did not enroll for the first trimester, 33 students without admission grade, and 150 students without university grade information (students who first enrolled in 2015 are in this group) were removed and finally 667 cases were remained. Two outcome categories are defined; one is dropout and consists of students who enroll in the first year but do not show up in the second year, the other one is underperformance and consists of students who fail 4 or more of the 12 subjects offered in the first year. Out of 667 cases, 286 students drop out and an additional 62 students underperform.

## 3.1 Per-Group Analysis

The average (base) risk rates of different groups are shown on Table 1. Foreign students have more risks compared to nationals, and the risk of students with lower admission grades is higher than the risk of students with higher admission grades. Naturally, students who fail more subjects and/or who have to take re-sit exams exhibit more risk than their counterparts. There have been two study programs for the total of 60 credits in the first year; plan A (older) with 10 courses and plan B (newer) including 12 courses. In the newer plan, with the aim of improving learning process, there is a course reorganization so that students can experience their first programming course in the first trimester and as can be seen, this change caused lower dropout and dropout/underperformance compared to the older plan.

Table 1: Per-group risk rates. Groups having 10 percentage points or more of risk compared to their counterparts are marked with an asterisk (*).

| Group | Size | Risk of Dropout | Risk of Dropout or Underperformance |
|---|---|---|---|
| Female | 9% | 41% | 54% |
| Male | 91% | 43% | 52% |
| Nationals[2] | 88% | 41% | 50% |
| Foreigners | 12% | 58% * | 69% * |
| Age ≤ 19 (median age) | 55% | 44% | 55% |
| Age > 19 | 45% | 41% | 48% |
| High school in same state ("in-State") | 76% | 44% | 53% |
| High school in another state ("out-of-State") | 24% | 41% | 49% |
| Public high school | 42% | 44% | 55% |
| Non-public high school | 58% | 42% | 50% |
| Avg. admission grade ≤ median | 50% | 49% * | 59% * |
| Avg. admission grade > median | 50% | 37% | 45% |
| Exam retake (at least once in first year) | 87% | 47% * | 58% * |
| No exam retake | 13% | 13% | 13% |
| Course failure (at least once in first year) | 85% | 47% * | 58% * |
| No course failure | 15% | 17% | 17% |
| Plan A (older) | 74% | 46% * | 53% |
| Plan B (newer) | 26% | 35% | 50% |
| Passed credits ratio[3] ≤ median | 50% | 70% * | 84% * |
| Passed credits ratio > median | 50% | 16% | 21% |

---

[2] For the purposes of this work, these are students who were born and are resident in the country.

[3] Number of credits passed over total credits during the first year

## 4 METHODOLOGY

We consider two predictive tasks: predicting dropout and predicting dropout or underperformance.

### 4.1 ML-based Models

According to the two ground truths (dropout, and dropout or underperformance), separate ML models are created. The feature set for the models consists of demographics (gender, age, and nationality), high school type and location, and average admission grade. Different ML algorithms: logistic regression, multi-layer perceptron (MLP), and support vector machines (SVM) are used to predict dropout risks. ML models are trained using cases enrolled between 2009 to 2013 (409 cases), then tested on students enrolled in 2014, 2016 and 2017 (258 cases). To mitigate the gender imbalance (only 9% of students are women), we use the SMOTE[4] algorithm (Chawla, 2002). We only apply SMOTE on the training set and keep the original class distributions in the test set to ensure valid results.

### 4.2 Algorithmic Fairness

Parity in the error rates of different groups ("equalized odds") is a well-established method to mitigate algorithmic discrimination in automatic classification (Hardt, 2016; Zafar, 2017; Woodworth, 2017). At the same time, we want to maintain model calibration (Dieterich, 2016; Berk R. a., 2018), as otherwise the same risk estimate carries different meanings and cannot be interpreted equally for different groups. Hence, a relaxation method (Pleiss, 2017) is used in this paper which seeks to satisfy equalized odds or parity in the error rates while preserving calibration. In most cases, calibration and equalized odds are mutually incompatible goals (Chouldechova A., 2017; Kleinberg, 2016), so in this method it is sought to minimize only a single error disparity across groups while maintaining calibration probability estimates.

If variable $x$ represents a student's features vector, $y$ indicates whether or not the student drops out, $G_1, G_2$ are the two different groups, and $h_1, h_2$ are binary classifiers which classify samples from $G_1, G_2$ respectively, Generalized False Positive Rate (GFPR) and Generalized False Negative Rate (GFNR) are defined as follows (Pleiss, 2017): the GFPR of classifier $h_t$ for group $G_t$ is $c_{fp}(h_t) = \mathbb{E}_{(x,y) \sim G_t}[h_t(x)|y = 0]$. This is the average probability of dropout that the classifier estimates for students who do not drop out. Conversely, the GFNR of classifier $h_t$ is $c_{fn}(h_t) = \mathbb{E}_{(x,y) \sim G_t}[(1 - h_t(x))|y = 1]$. So the two classifiers $h_1, h_2$ show probabilistic equalized odds across groups $G_1, G_2$ if $c_{fp}(h_1) = c_{fp}(h_2)$ and $c_{fn}(h_1) = c_{fn}(h_2)$. Classifier $h_t$ is said to be well-calibrated if $\forall p \in [0, 1], \mathbf{P}_{(x,y) \sim G_t}[y = 1|h_t(x) = p] = p$. To prevent the probability scores from carrying group-specific information, both classifiers $h_1, h_2$ are also calibrated with respect to groups $G_1, G_2$ (Berk R. a., 2018; Dieterich, 2016).

---

[4] Synthetic Minority Oversampling Technique

# 5 RESULTS

## 5.1 Effectiveness Evaluation

The best results for both dropout risk predictions were obtained using a Multi-Layer Perceptron (MLP). We used a single hidden layer having 100 neurons. The other models are omitted for brevity. Results in terms of the AUC-ROC, GFNR, GFPR, and F-score (the harmonic mean of precision and recall, which unlike the other metrics, requires to establish an optimal cut-off for classification) are presented in Table 2. According to the results, the models lead to good performance in terms of AUC and F-score in both prediction tasks. With a little information at the time of students' enrollment, these models show good AUC in comparison to previous work (Aulck, 2016; Nagy, 2018) which showed AUC in the order of 0.62-0.81. Also, comparing calibrated and non-calibrated predictions we can see that calibrated model leads to lower GFNR and non-calibrated results in lower GFPR.

## 5.2 Algorithmic Fairness Evaluation

The results for the analysis of algorithmic fairness are shown on the left side of Table 3. In dropout prediction, we can observe accuracy equity (less than 20% discrepancy) in terms of AUC in both models, even if results are slightly more accurate for male students. AUC is also higher for students with lower admission grades compared to their counterparts. In the calibrated model, males, foreigners, and lower admission grade students experience lower GFNR compared to their counterparts. However, non-calibrated model shows fairer results for GFNR along these groups. Regarding GFPR, there can be seen more false positive errors (higher risk scores for students who do not dropout or underperform) for males compared to females, students of out-of-State high schools than in-State high schools, and lower admission grade students compared to their counterparts in the non-calibrated model. In the calibrated model, this metric shows more errors for foreigners and for lower admission grade students compared to their counterparts.

Similar results are shown for predicting dropout or underperformance. In terms of AUC, MLP shows equity (less than 20% discrepancy) across groups except for more accuracy for students from in-State high schools. In the calibrated model, higher AUC can be observed in nationals compared to foreigners and higher admission grade students. Also, both models show parity across all groups in terms of GFNR except for students with lower admission grade who experience lower errors compared to their counterparts, however, non-calibrated model shows lower discrimination to this groups compared to the calibrated one. In terms of GFPR, we can see more errors of the model for foreigners than nationals, out-of-State high school than in-State high school students, males than females, and cases with lower admission grades compared to their counterparts. In the calibrated model, this metric also shows more error for foreigners than national and students with lower admission grade compared to their counterparts, but it reveals more errors for females than males.

Table 2: Effectiveness of models in risk prediction.

| Risk | Dropout | | | | Dropout or Underperformance | | | |
|---|---|---|---|---|---|---|---|---|
| Model | AUC | GFNR | GFPR | F-score | AUC | GFNR | GFPR | F-score |
| MLP | 0.77 | 0.73 | 0.19 | 0.76 | 0.78 | 0.69 | 0.19 | 0.83 |
| MLP calibrated | 0.77 | 0.36 | 0.42 | 0.76 | 0.78 | 0.27 | 0.49 | 0.83 |

# 6      EQUALIZED ODDS AND CALIBRATION

In this section, parity is sought along groups in terms of two fairness metrics. For this purpose, the method introduced by Pleiss et al. (Pleiss, 2017) is used, which seeks parity in Generalized False Positive Rate (GFPR) or Generalized False Negative Rate (GFNR) while preserving calibration. In both prediction tasks, the models before mitigation exhibit in general better parity in terms of AUC and GFNR and more inequality in terms of GFPR. The results after bias mitigation are presented in the right side of the Table 3. By comparing the results before and after GFPR bias mitigation in dropout we can see that the disparity in GFPR has decreased in the order of 0.03-0.71 in MLP and 0.02-0.30 in MLP calibrated across all groups. Also, comparing the result before and after GFPR bias mitigation in dropout or underperformance show that bias in MLP and MLP calibrated models has been respectively reduced by the order of 0.08-1.15 and 0.14-0.59 across all groups.

# 7      CONCLUSIONS AND RECOMMENDATIONS

The effectiveness and fairness of Machine Learning (ML) models in the early prediction of university dropout and underperformance was evaluated. Using only information at the time of enrollment, calibrated ML models were created with AUC of 0.77 and 0.78 which can help reliably identify students at risk to trigger interventions that can help increase their success and ultimately reduce social and economic costs. When introducing ML models, improvements in accuracy need to be carefully contrasted with potential algorithmic discrimination. Thus, we evaluated the algorithmic fairness of the ML models in terms of AUC and error (GFNR and GFPR) across five groups defined by nationality, gender, high school type and location, and admission grade. According to the results, our modeling has parity in terms of AUC and GFNR but disparities in GFPR. These disparities in GFPR are larger among groups defined by admission grade, and the bias is against students with lower admission grades. The predicted probability of dropout for the students of this sub group who do not actually drop out is larger than that of their counterparts (students of higher admission grade sub group). Using a relaxation method (Pleiss, 2017), we tried to obtain parity in GFPR while preserving calibration. By maintaining the calibration among subgroups, we prevent the probability scores from needing group-dependent interpretation. The results after bias mitigation show that GFPR ratio in both dropout and dropout or underperformance predictions has been changed to a perfect value close to 1 across most of the groups. This bias mitigation also caused better parities in other metrics (AUC and GFNR) along majority of the groups compared to the non-mitigated model. Studying algorithmic discrimination means addressing unfair decisions not only to the identification of students that would require preventive mentoring programs, but also to the identification of potentially successful students that would benefit from e.g. additional educational opportunities or to the formulation of pedagogical interventions related to changes in the study plans or in pedagogical methods suiting specific students' profiles.

In terms of contributions to learning analytics, in addition to creating ML models for dropout and underperformance that exhibit high accuracy, we evaluated algorithmic fairness of the models across different groups in terms of several metrics and applied a bias mitigation method to set parity for subgroups with unfair results. For the students at high risk of dropout or underperformance, different interventions can be considered such as tutoring, counselling and mentoring. A suggested beneficial intervention (Lowis, 2008) is interviewing with the students in informal discussion and asking for their perceptions and experiences at the university which can help with the planning process for their subsequent academic years. Also, a preventive mentoring program (Larose, 2011) showed high levels of motivation and more positive career decision profiles for the newcomer students who participated in bimonthly meetings with students completing their undergraduate degree. Both require early prediction models with equity among groups, which the methods we have described can provide in a real-world setting.

## ACKNOWLEDGMENTS

This work has been partially supported by the HUMAINT programme (Human Behavior and Machine Intelligence), Centre for Advanced Studies, Joint Research Centre, European Commission. The project leading to these results have received funding from "la Caixa" Foundation (ID 100010434), under the agreement LCF/PR/PR16/51110009. We also acknowledge the support of ICREA Academia.

Table 3: Effectiveness (AUC) and fairness (GFPR and GFNR ratios) of models for the two risk prediction tasks, before and after bias mitigation. Values in boldface should, ideally, be close to 1.0 to indicate perfect equity among groups.

| | Before bias mitigation | | | | | | | | | | | | After bias mitigation | | | | | | | | | | | |
|---|---|---|---|---|---|---|---|---|---|---|---|---|---|---|---|---|---|---|---|---|---|---|---|---|
| Risk | Dropout | | | | | | Dropout or Underperformance | | | | | | Dropout | | | | | | Dropout or Underperformance | | | | | |
| Model | MLP | | | MLP calibrated | | | MLP | | | MLP calibrated | | | Equalized GFPR MLP | | | Equalized GFPR MLP calibrated | | | Equalized GFPR MLP | | | Equalized GFPR MLP calibrated | | |
| Group \ Metric | AUC | GFNR | GFPR | AUC | GFNR | GFPR | AUC | GFNR | GFPR | AUC | GFNR | GFPR | AUC | GFNR | GFPR | AUC | GFNR | GFPR | AUC | GFNR | GFPR | AUC | GFNR | GFPR |
| Nationals | 0.77 | 0.74 | 0.19 | 0.76 | 0.34 | 0.42 | 0.77 | 0.72 | 0.19 | 0.81 | 0.23 | 0.45 | 0.77 | 0.74 | 0.19 | 0.50 | 0.46 | 0.54 | 0.64 | 0.68 | 0.28 | 0.50 | 0.35 | 0.65 |
| Foreigners | 0.82 | 0.73 | 0.17 | 0.69 | 0.26 | 0.61 | 0.67 | 0.66 | 0.25 | 0.55 | 0.21 | 0.67 | 0.82 | 0.72 | 0.17 | 0.69 | 0.26 | 0.61 | 0.67 | 0.66 | 0.25 | 0.55 | 0.21 | 0.67 |
| $\frac{\text{Nationals}}{\text{Foreigners}}$ (Ratio) | **0.93** | **1.01** | **1.13** | **1.10** | **1.33** | **0.68** | **1.15** | **1.08** | **0.76** | **1.45** | **1.07** | **0.67** | **0.93** | **1.03** | **1.13** | **0.73** | **1.79** | **0.88** | **0.95** | **1.03** | **1.14** | **0.90** | **1.63** | **0.98** |
| State_Highschool | 0.79 | 0.74 | 0.18 | 0.74 | 0.35 | 0.46 | 0.79 | 0.72 | 0.17 | 0.78 | 0.22 | 0.47 | 0.65 | 0.71 | 0.25 | 0.74 | 0.35 | 0.46 | 0.62 | 0.62 | 0.32 | 0.72 | 0.27 | 0.54 |
| NonState_Highschool | 0.71 | 0.70 | 0.23 | 0.77 | 0.31 | 0.43 | 0.59 | 0.67 | 0.30 | 0.74 | 0.25 | 0.54 | 0.71 | 0.70 | 0.23 | 0.78 | 0.30 | 0.43 | 0.59 | 0.67 | 0.30 | 0.74 | 0.25 | 0.54 |
| $\frac{\text{State\_Highschool}}{\text{NonState\_Highschool}}$ (Ratio) | **1.11** | **1.06** | **0.77** | **0.96** | **1.12** | **1.07** | **1.35** | **1.08** | **0.57** | **1.05** | **0.88** | **0.87** | **0.92** | **1.02** | **1.07** | **0.94** | **1.13** | **1.09** | **1.06** | **0.93** | **1.08** | **0.97** | **1.06** | **0.99** |
| Pub_Highschool | 0.82 | 0.74 | 0.17 | 0.77 | 0.36 | 0.39 | 0.79 | 0.71 | 0.18 | 0.78 | 0.22 | 0.44 | 0.80 | 0.71 | 0.20 | 0.63 | 0.41 | 0.50 | 0.81 | 0.68 | 0.18 | 0.75 | 0.24 | 0.51 |
| NonPub_Highschool | 0.72 | 0.73 | 0.21 | 0.72 | 0.31 | 0.47 | 0.72 | 0.70 | 0.21 | 0.75 | 0.20 | 0.49 | 0.72 | 0.73 | 0.21 | 0.72 | 0.31 | 0.47 | 0.72 | 0.70 | 0.21 | 0.75 | 0.20 | 0.49 |
| $\frac{\text{Pub\_Highschool}}{\text{NonPub\_Highschool}}$ (Ratio) | **1.14** | **1.01** | **0.85** | **1.07** | **1.19** | **0.82** | **1.09** | **1.01** | **0.88** | **1.03** | **1.07** | **0.91** | **1.10** | **0.96** | **0.96** | **0.88** | **1.33** | **1.05** | **1.11** | **0.97** | **0.88** | **0.99** | **1.18** | **1.05** |
| Low_AdmissionGrade | 0.69 | 0.70 | 0.27 | 0.67 | 0.24 | 0.58 | 0.66 | 0.65 | 0.33 | 0.55 | 0.15 | 0.77 | 0.69 | 0.70 | 0.27 | 0.67 | 0.24 | 0.58 | 0.66 | 0.65 | 0.33 | 0.55 | 0.15 | 0.77 |
| High_AdmissionGrade | 0.52 | 0.84 | 0.16 | 0.71 | 0.50 | 0.37 | 0.64 | 0.82 | 0.15 | 0.74 | 0.41 | 0.40 | 0.52 | 0.73 | 0.27 | 0.50 | 0.57 | 0.43 | 0.49 | 0.67 | 0.34 | 0.50 | 0.51 | 0.49 |
| $\frac{\text{Low\_AdmissionGrade}}{\text{High\_AdmissionGrade}}$ (Ratio) | **1.32** | **0.83** | **1.68** | **0.95** | **0.49** | **1.59** | **1.03** | **0.79** | **2.16** | **0.74** | **0.37** | **1.95** | **1.32** | **0.96** | **1.01** | **1.35** | **0.42** | **1.36** | **1.33** | **0.97** | **0.97** | **1.10** | **0.30** | **1.57** |
| Male | 0.78 | 0.73 | 0.20 | 0.75 | 0.33 | 0.45 | 0.77 | 0.70 | 0.20 | 0.78 | 0.23 | 0.47 | 0.78 | 0.73 | 0.20 | 0.75 | 0.33 | 0.45 | 0.77 | 0.70 | 0.20 | 0.51 | 0.34 | 0.65 |
| Female | 0.62 | 0.84 | 0.13 | 0.57 | 0.51 | 0.44 | 0.67 | 0.80 | 0.15 | 0.71 | 0.23 | 0.64 | 0.62 | 0.76 | 0.20 | 0.67 | 0.55 | 0.45 | 0.56 | 0.80 | 0.22 | 0.71 | 0.23 | 0.64 |
| $\frac{\text{Male}}{\text{Female}}$ (Ratio) | **1.26** | **0.87** | **1.51** | **1.33** | **0.66** | **1.02** | **1.15** | **0.87** | **1.37** | **1.10** | **0.99** | **0.73** | **1.26** | **0.96** | **1.00** | **1.13** | **0.61** | **1.01** | **1.38** | **0.87** | **0.91** | **0.72** | **1.45** | **1.02** |